\documentclass{article}

\usepackage{arxiv}

\usepackage[utf8]{inputenc} % allow utf-8 input
\usepackage[T1]{fontenc}    % use 8-bit T1 fonts
\usepackage{hyperref}       % hyperlinks
\usepackage{url}            % simple URL typesetting
\usepackage{booktabs}       % professional-quality tables
\usepackage{amsfonts}       % blackboard math symbols
\usepackage{nicefrac}       % compact symbols for 1/2, etc.
\usepackage{microtype}      % microtypography
\usepackage{lipsum}
\usepackage{graphicx}
\graphicspath{ {./images/} }
\usepackage{amsmath}

\usepackage{booktabs,xcolor,colortbl,multirow,rotating,bigstrut} % for Excel2LaTeX tables
\usepackage{adjustbox}

\usepackage[nameinlink]{cleveref}

\title{Frugal Algorithm Selection} 
\author{
  Erdem Kuş, Özgür Akgün, Nguyen Dang, Ian Miguel \\
  School of Computer Science \\
  University of St Andrews \\
  St Andrews\\
  \texttt{\{ek232, oa86, nttd, ijm\}@st-andrews.ac.uk} 
}

\begin{document}

\maketitle

\begin{abstract}

% 157 words
% When solving decision and optimisation problems, many competing algorithms (by which we include both model and solver choices) have complementary strengths. Typically, there is no single algorithm that works well for all instances of a problem. Automated algorithm selection has been shown to work very well for choosing a suitable algorithm for a given instance. However, the cost of training an automatic algorithm selection system can be prohibitively large due to running candidate algorithms on a representative set of training instances. In this work, we explore reducing this cost by choosing a subset of the training instances on which to train. We approach this problem in 3 ways: using active learning-based methods to decide based on prediction uncertainty, augmenting the algorithm predictors with a timeout predictor, and collecting training data using a progressively increasing timeout. We evaluate combinations of these approaches on 6 datasets from ASLib and present the reduction in training time achieved by each option.

% 148 words
When solving decision and optimisation problems, many competing algorithms (model and solver choices) have complementary strengths. Typically, there is no single algorithm that works well for all instances of a problem. Automated algorithm selection has been shown to work very well for choosing a suitable algorithm for a given instance. However, the cost of training can be prohibitively large due to running candidate algorithms on a representative set of training instances. In this work, we explore reducing this cost by choosing a subset of the training instances on which to train. We approach this problem in three ways: using active learning to decide based on prediction uncertainty, augmenting the algorithm predictors with a timeout predictor, and collecting training data using a progressively increasing timeout. We evaluate combinations of these approaches on six datasets from ASLib and present the reduction in labelling cost achieved by each option.

\end{abstract}

\section{Introduction}

Solving combinatorial optimisation problems is a challenging task, where often multiple approaches compete to offer the most effective solution. In many cases, these problems require large amounts of computational resource. Typically, different algorithms (model and solver choices) are better suited for different problem instances and new potential approaches are continuously developed. Identifying the most suitable algorithm for a particular problem instance has the potential to provide significant efficiency gains.

Machine learning (ML) has emerged as a powerful tool for automating algorithm selection, effectively learning to predict which algorithm is most likely to perform well on a given problem instance. By analysing the features of various problem instances and the corresponding performance of different algorithms, ML models can be trained to recommend the most suitable algorithm for a specific problem~\cite{kotthoff2016algorithm,kerschke2019automated}. This approach has proven highly effective, often leading to significant improvements in computational efficiency and overall problem-solving capability (e.g.,~\cite{xu2008satzilla,xu2012satzilla2012,ansotegui2016maxsat}). Despite the success of ML-based algorithm selection techniques, there is a notable drawback: the high computational cost associated with running candidate algorithms on a large and representative set of training instances. In order to generate a robust model, it is necessary to evaluate the performance of each algorithm on numerous problem instances. However, this process can be time-consuming and resource-intensive, limiting the scalability and practicality of current algorithm selection methods.

% https://en.wikipedia.org/wiki/Active_learning_(machine_learning) 
Herein we propose to select instances for training AS models iteratively. This setting is called Active Learning (AL) \cite{Cohn2010}, and is popular for cost-effective training in ML. We focus on AS scenarios that aim at optimising the runtime of the predicted algorithm on a specific instance, a common setting in CP and SAT domains. In these scenarios, each run of an algorithm on a problem instance is limited by a typically large cutoff time. Full information about performance data of timeout cases is expensive to collect but not necessarily useful.

In a typical AL scenario, the {\em labelling cost} (in our context the cost of running an algorithm to solve a selected instance) is assumed to be uniform. Our situation is further complicated by the fact that, within the cutoff, there is typically significant variance in the cost of running an algorithm. Therefore, we propose two strategies within the active learning setting to reduce labelling cost, thus improving learning efficiency. We evaluate the proposed strategies in combination with two methods for selecting new instances in the active learning setting: an uncertainty-based and a naive random selection method (see~\Cref{sec:instance_selection} for details). Experimental results on six scenarios from ASLib~\cite{BISCHL201641}, the standard benchmarking library for algorithm selection, are presented, showing the effectiveness of applying active learning setting to the AS context: in most cases, we can achieve 100\% of the predictive power of an AS method that uses all of the training data, while requiring less than 60\% of total labelling cost. This is further reduced to as low as 10\% of the labelling cost in some scenarios, thanks to a combination of using a timeout predictor and dynamic timeouts.

%We select instances for training AS models in an iterative fashion. This setting is called Active Learning (AL) \cite{Cohn2010}, and popular setting for cost-effective training in ML. 

%We consider instance selection approaches: AL-based and random selection...

%Information about timeout cases is expensive to collect but not always useful. To aid the selection process, we propose two strategies: timeout predictor and dynamic timeout...

% The primary contributions of our work are: 
% \begin{enumerate}
% \item Novel use of timeout predictors and dynamic timeouts to mitigate annotation cost variance.
% \item Thorough empirical analysis for evaluating the new and existing ideas.
% \item Discovery that timeout aware active learning performs best across our benchmarks.
% \end{enumerate}

% \nguyen{another attempt}
The primary contributions of our work are: 
\begin{enumerate}
\item Novel use of timeout predictors and dynamic timeouts for cost-effective algorithm selection.
\item A thorough empirical evaluation across six benchmarks from ASLib, validating the effectiveness of the proposed timeout aware active learning approach.
\end{enumerate}

\section{Background}

This section provides background for our work, covering three key areas: Automated Algorithm Selection, where the primary challenge is to optimally select an algorithm from a portfolio based on specific instance characteristics and desired performance outcomes; ASLib, a pivotal resource for benchmarking algorithm selection; and Active Learning, a method for organising the training process by strategically choosing informative data points to label.

\subsection{Automated Algorithm Selection\label{sec:automated_algorithm_selection}}

The automated algorithm selection (AS) problem can be defined as follows. Given a portfolio of $n$ algorithms $\mathcal{A}=\{a_1, a_2, ..., a_n\}$ with \emph{complementary} strengths, an instance distribution $\mathcal{D_I}$ where each instance $i \sim \mathcal{D_I}$ is described as a feature vector $v(i)$, a metric $c(i,a_k)$ that measures the performance of algorithm $a_k \in \mathcal{A}$ on instance $i$, the AS problem involves building an automated selector $f(v(i))$ to select the best algorithm $a \in \mathcal{A}$ for an instance $i$. More formally, we want to find $f$ such that $E_{i \sim \mathcal{D_I}} [ c(i, f(v(i)) ]$ is optimised. There is sometimes an extra cost associated with the extraction of instance features, which must be taken into account in our optimisation objective. In practice, we are often given in advance a set of training instances, their feature vectors, the cost of extracting such features, and the performance of each algorithm in the portfolio on each given instances. This data is used for training a machine learning model to predict the best algorithm for an unseen instance.

The first AS problem was described in 1976 by Rice~\cite{rice1976algorithm}. Over the last two decades, several AS approaches have been proposed and the applications of such techniques have provided a significant boost in state-of-the-art performance across a wide range of computational areas, such as Constraint Programming~\cite{o2008using}, propositional satisfiability (SAT) solving~\cite{xu2008satzilla,xu2012satzilla2012}, AI planning~\cite{vallati2014asap,rizzini2015portfolio,rizzini2017static}, and combinatorial optimisation~\cite{kerschke2018leveraging}. 
A wide range of machine learning-based techniques were adopted in those AS approaches. Some examples include empirical performance models~\cite{xu2008satzilla}, k-nearest neighbours~\cite{collautti2013snnap}, clustering-based methods~\cite{kadioglu2010isac,ansotegui2016maxsat}, and cost-sensitive pairwise classification approaches~\cite{xu2012satzilla2012}. In addition to building a machine learning model to predict the best performing algorithm, modern AS systems often adopt an extra component called a \emph{pre-solving schedule}~\cite{xu2012satzilla2012,hoos2014claspfolio,lindauer2015autofolio,gonard2019algorithm}, a static schedule of algorithms run for a small amount of time before (expensive) feature extraction and algorithm selection are conducted. Another related approach is algorithm scheduling, where instead of selecting a single algorithm for given instance, we build a \emph{schedule} of algorithms~\cite{amadini2014sunny,amadini2015sunny,liu2021sunny}. For detailed overviews of AS approaches and their applications, we refer to~\cite{kerschke2019automated,kotthoff2016algorithm}.

\subsection{The Algorithm Selection Library (ASLib)\label{sec:aslib}}

The Algorithm Selection Library~\cite{BISCHL201641} (ASLib) is a widely-used benchmarking library for automated algorithm selection. The library provides a standard format for representing algorithm selection scenarios across a wide range of algorithms and problems, and currently consists of $44$ datasets from multiple application domains, including SAT, Quantified Boolean Formula (QBF), Maximum Satisfiability (MAX-SAT), Constraint Satisfaction Problems, Answer Set Programming (ASP), and combinatorial optimisation problems. In those scenarios, the performance metric is either the algorithm running time (for solving a given instance) or the solution quality obtained within a time limit.

The process of data labelling, i.e. obtaining the target output labels for all training data points, is a critical task in the realm of machine learning, as it is an essential component for building effective models. However, this process is not free of cost, and it can be a time-intensive endeavour, often requiring more resources than training the machine learning model itself. In context of automated algorithm selection, the cost of collecting the performance data can be substantial. As an example, several scenarios in the ASLib require more than 100 CPU days to collect the full performance data on the given instance sets (some may require up to $3$ CPU years). Our hypothesis is that we can significantly reduce this cost by only collecting partial information about the performance data without sacrificing too much in terms of algorithm selection quality.

\subsection{Active Learning for querying informative instances\label{sec:active_learning}}

Active learning is a methodology that maintains machine learning accuracy with fewer labelled data points by allowing incremental labelling of training data~\cite{Settles2009ActiveLL, 6747346}. This iterative process includes training a model on a small set of labelled data, using the model to identify and query the most uncertain unlabelled data points, obtaining labels for these queried points, and then retraining the model with the newly labelled data.

% Active learning is a machine learning methodology that involves human interaction to achieve higher accuracy with less labelled data where data labelling costs are expensive. This process involves an iterative cycle that takes place until the model achieves the desired predictive performance i) training a machine learning model based on the labelled data; ii) predicting from unlabelled data and \emph{querying} data which it makes most uncertain predictions for labelling iii) human (or other) annotator labelling data has been queried by model iv) model is retrained with updated labelled data

There are multiple methods for querying in active learning~\cite{5272205, WANG2015426, 10.5555/2999611.2999774, 10.5555/2999325.2999397}.
% related to data sampling querying the data for labelling in active learning. In our study, pool-based sampling was used for data sampling and uncertainty-based sampling was used for query, which are the most commonly used methods for active learning and the most suitable methods for our problem at the same time.
In our study, we use pool-based sampling (we maintain a list of candidate training data points that can be queried) and we use two methods for querying from this pool: uncertainty-based instance selection and random instance selection. These are widely regarded as effective methods for active learning, particularly suitable for our context. See \Cref{app:a} for more information about the behaviour of alternative query methods.

\section{Frugal Algorithm Selection}

This section explains our approach to automated algorithm selection and the three opportunities we have identified for reducing the amount of resources needed for training.

\textbf{Machine learning setup -- } We use a collection of binary classifiers, each designed to compare pairs of algorithms to determine which is faster for a given instance. Using these binary classifiers, we implement a multi-class classification method through a voting system, where each algorithm receives a vote each time it is predicted to be the quicker option compared to another. The algorithm that accumulates the most votes across all comparisons is then chosen as the best option~\cite{bishop2006pattern}. This setup is advantageous for our goal of minimising resource usages, as it allows selective training of classifiers on specific subsets of instances.

\textbf{Passive learning -- } For each binary classifier, we train a random forest predictor using the entire training set. For all but one of the datasets (MAXSAT19-UCMS), passive learning performs better than the single best algorithm (always using the algorithm that performed the best for the training set) and for MAXSAT19-UCMS passive learning performs slightly worse than the single best option (details can be found in \Cref{app:d}).

\textbf{Frugal methods -- } We explore three configuration options, each offering two alternatives, to create a range of strategies for frugal algorithm selection. The first configuration option is \underline{instance selection}, which involves comparing uncertainty-based selection (focusing on potentially informative instances) with random selection. The second configuration option is whether we use \underline{timeout predictors} and the third configuration option is whether we use \underline{dynamic timeouts}.
% \textbf{Frugal methods -- } We evaluate 3 configuration settings, each with 2 possible options to frugal algorithm selection. First configuration is regarding how we select instances, the options here are uncertainty-based selection and random selection. The second configuration is whether we use a timeout predictor for each algorithm in addition to the binary classifiers. The third option is dynamic timeouts: whether we always run an algorithm on an instance to completion (with an ultimate time limit of one hour) or we start with a short timeout value and increase the timeout value iteratively.
In the rest of this section we explain these three configuration options. In \Cref{sec:experimental} we empirically evaluate all 8 combinations of these configuration settings.

% %%%%%%

% Also, the following is a good argument for us:
 
% Active learning is based on informativeness (via prediction uncertainty) for choosing training samples. Standard methods of doing this do not differentiate between cost of collecting training samples: i.e. they assume each training sample costs the same amount. In our setting of algorithm selection, however, some samples are significantly cheaper than others.
 
% We want to balance 'bang for the buck': maximise how much information is gained per time spent. Something like cost-aware active learning, or active learning with non-uniform cost.
 
% Hypothesis 1: Time out predictors can help.
 
% Hypothesis 2: Progressively increasing the timeout can help.

% %%%%%%

% Argument for why RQ also works even though it doesn't care about informativeness.
 
% Majority of the cost goes to instances that time out. Eliminating these through TO and DT means informativeness is less important than avoiding high cost instances.
 
% Future work: a hybrid setup where we start with RQ (it's cheaper in terms of feature cost) and we switch to AL later in the process when RQ reaches a plateau. Similar to a cooling schedule in simulated annealing.

% %%%

\subsection{Selecting Instances: Uncertainty-based vs random\label{sec:instance_selection}}

In our frugal algorithm selection methods, we begin by training all machine learning classifiers on a small number of randomly selected instances. The remaining instances in the training set are kept in a pool of candidates. For each classifier we maintain a separate pool of candidates: this allows us to run an instance on a subset of the algorithms instead of necessarily running it on all algorithm options. This flexibility can be particularly useful when some algorithms tend to timeout very often and hence take up a lot of resources unnecessarily.

% should this be (un)annotated rather than (un)labelled throughout?
At each step of our algorithm, we select \textit{N} samples from the available unlabelled set of instances. One option for the selection process is employing an uncertainty-based query strategy. Based on an ``informativeness'' measure, this strategy aims to prioritise the instances that are most likely to yield valuable insights when annotated. We perform uncertainty-based querying by using the predictive model we have partially trained so far. We predict a class for each candidate instance in the pool. The machine learning predictor returns a confidence level in addition to a class prediction. We then query the instances where the predictor is least confident. Uncertainty-based querying is based on the premise that, by focusing on the data points where the model’s predictions are least confident, the model is expected to learn from the most uncertain data points. We also allow different machine learning models to make different numbers of queries, based on the confidence levels. For each predictive model we create a table of requested data points and the associated confidence level. We then combine these tables into a single table, sort the table by confidence level and select top \textit{N} requests. This approach enables predictive models with a high level of uncertainty to query more instances in comparison to those with very low levels of uncertainty.

Uncertainty-based querying can be expensive because it requires feature extraction for all instances in the candidate pool at the start of the procedure. It also only considers informativeness without taking labelling cost into account. To evaluate whether uncertainty-based querying is effective in our work, we compare it with a random query order.

\subsection{Timeout Predictor}

Instances that time out with a given algorithm are particularly costly in automated algorithm selection. This is partly because if an instance cannot be solved by two algorithms within the timeout, we spend considerable time running these algorithms but gain no new information. Furthermore, when one algorithm solves an instance quickly and another times out, we gain no additional information by allowing the slower algorithm to run to completion. The binary classifiers are provided information only about which algorithm is faster.
% Excised because it seems to be repetitious
% Thus, significant time may be wasted with minimal or no information gain when algorithms time out.

%\nguyen{clarify how timeout predictor is used during the instance selection process: we use uncertainty measurement to sort the unlabelled data. After that, we apply timeout predictors. All pairs of (algorithm, instance) that is predicted as timeout will be pushed to the end of the selection queue.}

We enhance our base machine learning architecture of binary classifiers for all algorithm pairs with dedicated timeout predictors: additional random forest classifiers, one per algorithm, whose task is to predict whether an unseen instance will time out for a specific algorithm. The hypothesis is that training a timeout predictor is a simpler learning task, and this classifier can be trained without requiring additional data. We adjust our voting mechanism to take the timeout predictors into account. If an algorithm is predicted to time out, we exclude it from the options before calculating the votes using the pairwise predictors. An exception is in instances where all algorithms are predicted to time out; in such cases, we do not eliminate any of the options and continue to use pairwise predictors for the entire set.

\subsection{Dynamic Timeout\label{sec:DT}}

The dynamic timeout strategy begins with an initially defined timeout period and incrementally increases it up to a maximum of one hour. After the initial training phase in active learning, the algorithms selected for querying are executed on the selected data within the current time limit. An algorithm that fails to solve the example within this specified time is classified as a timeout for the active time limit. This approach is intended to minimise labelling costs by initially running instances with a short time limit. Hence, resources are not wasted on instances that both algorithms are likely to fail to solve in the early stages. In single timeout cases, where one of the two algorithms can solve the instance, samples can be labelled at a lower cost. The condition for increasing the timeout is determined based on the performance of the model predictions on the validation set. If the model predictions reach a plateau on the validation set, we increase the timeout at the specified rate.

When timeout predictors and dynamic timeouts are used in combination we train the timeout predictors with respect to the particular timeout value at a given moment.

% The primary goal of this approach is to minimise labelling costs for double timeout instances by running them within a short time interval. This strategy ensures that resources are not wasted on instances that both algorithms are likely to fail to solve within the early stages of active learning. In cases of single timeouts—where one of the two algorithms can solve the instance—samples can be labelled at a lower cost. 

\section{Experimental Results\label{sec:experimental}}

We evaluate the performance of all eight configurations of the frugal algorithm selection methods using six datasets from ASLib, chosen for their wide-ranging characteristics. These datasets include various problem-solving domains: one from Answer Set Programming (ASP), two from Constraint Programming (CP), two from propositional satisfiability (SAT), and one from Quantified Boolean Formula (QBF) solving. The datasets vary significantly in complexity and size, with between 2 to 11 algorithm options, 22 to 138 features, and 527 to 2024 instances. \Cref{app:c} presents detailed descriptive statistics of the selected datasets. 

\textbf{Experimental setup --} To evaluate our methodology we split each dataset into three subsets: training, validation, and test. We allocate 10\% of the instances to the test set and perform a 10-fold-cross-validation to ensure thorough evaluation, running each fold five times with different seeds to address randomness. An additional 10\% of the training set is used as the validation set, used to decide when to increase the timeout as discussed in \Cref{sec:DT}. At each step, we label 1\% of the training data.

Throughout, we multiply timeout values by 10 (the PAR10 measure). In all configurations, we remove features where more than 20\% of the instances have missing values and apply a median imputer to fill the remaining gaps.
% gini: https://towardsdatascience.com/gini-impurity-measure-dbd3878ead33
% bootstrapping: https://medium.com/@sly.of.zero/decision-trees-bootstrap-aggregating-and-bagging-8c6cf764e689
% for max_features, min_samples_split, scikit docs are good
We use a random forest classifier configured to use 100 estimators, using the Gini impurity measure to determine the best splits. Each tree is limited to using up to the square root of the number of features. The depth of the decision trees is practically unlimited (setting maximum depth to $2^{31}$), and we require at least two samples before splitting a node. We also turn on bootstrapping when sampling the data for building each decision tree. We have reached these settings through experimenting with the passive learning setup and used the same hyper-parameter settings throughout.

% for the record, hyperparameters are:
%         'n_estimators': 100,
%         'criterion': 'gini',
%         'max_features': 'sqrt',
%         'max_depth': 2**31,
%         'min_samples_split': 2,
%         'bootstrap' : True

We present our results in a series of plots designed to compare the configurations at different aggregation levels. All have the same structure. 
%The horizontal axis is the ratio of the total time taken to solve all instances using the predictions of the frugal method to that taken using the predictions of the passive learning method. 
The horizontal axis is the ratio between performance of frugal method and passive learning, where  performance is measured as the total runtime of the predicted algorithms on all test instances.
This metric serves as a proxy for predictive performance, indicating how closely the frugal method approaches the benchmark established by passive learning.
The vertical axis is the minimum amount of training data (as a ratio of the entire training set) required to achieve the performance indicated on the horizontal axis. The representation highlights the efficiency of the frugal method in terms of data utilisation. We plot the mean and a ribbon around the mean showing standard error.

%The source code used for our experiments, data files, and complete set of results can be found in the supplementary files repository (link anonymised for double-blind review): \url{https://anonymous.4open.science/r/CP2024-Frugal-6E4F}
All source code, data, experimental results and the appendix are available at \url{https://github.com/stacs-cp/CP2024-Frugal}.

\begin{figure}
    \centering
    \includegraphics[width=0.8\textwidth]{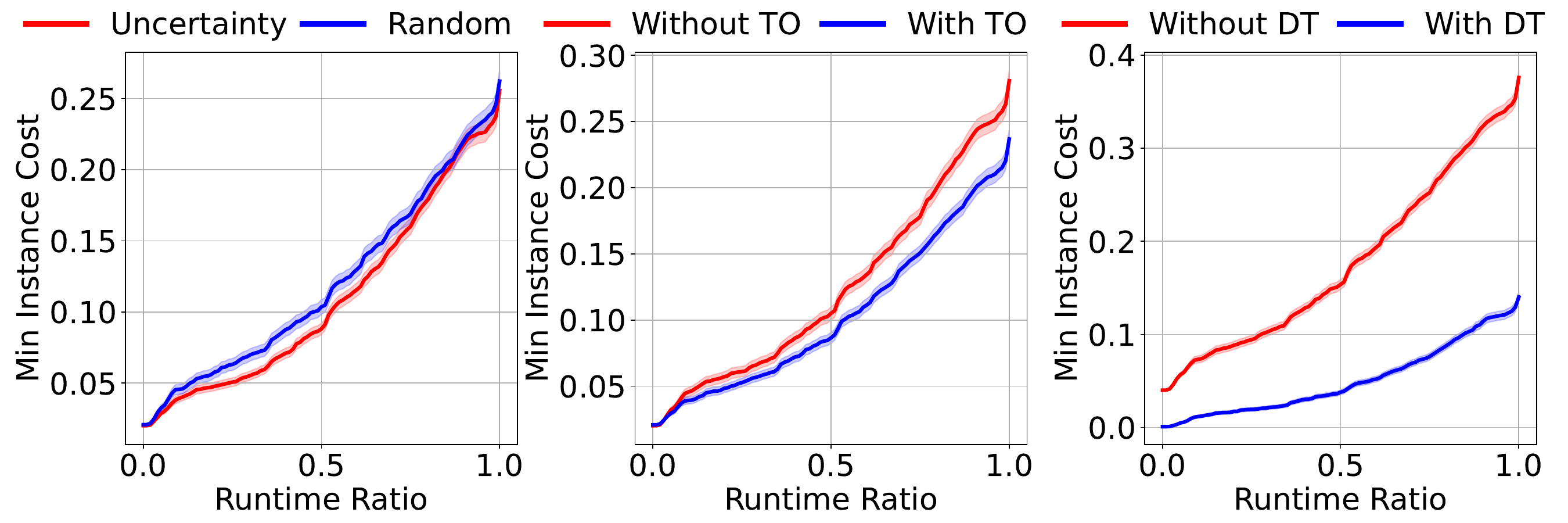}
    \caption{Results aggregated by configuration option. Instance selection (random vs uncertainty-based) does not make a big difference, timeout predictor (TO) improves runtime ratio slightly, dynamic timeout (DT) improves runtime ratio significantly.\label{fig:agg}\label{fig1}}
\end{figure}

\begin{figure}
    \centering
    \includegraphics[width=0.8\textwidth]{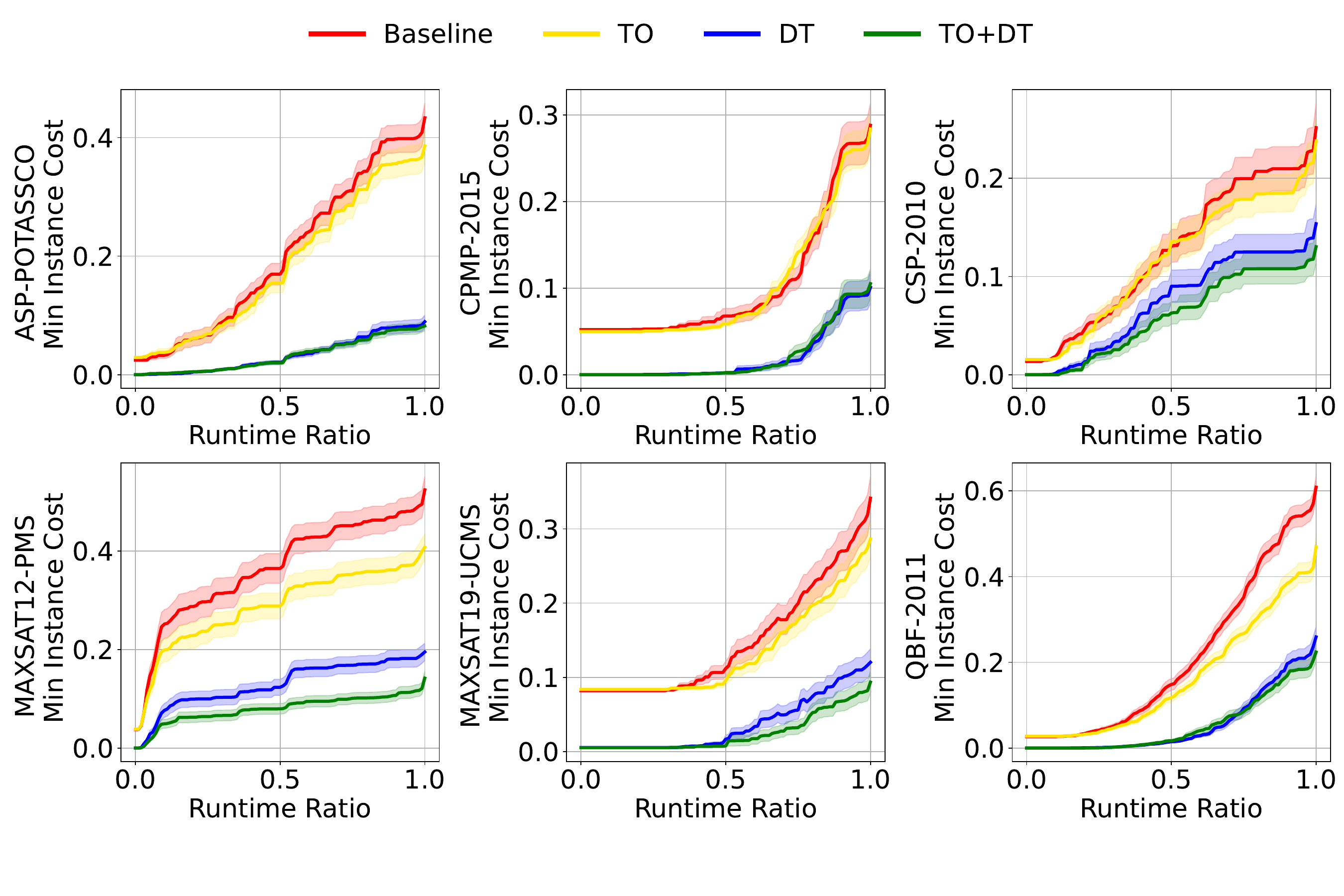}
    \vspace*{-8mm}
    \caption{Performance of different timeout configurations (aggregated by instance selection). Notably, the combination of timeout predictor and dynamic timeouts (TO+DT) and the standalone dynamic timeout option exhibit significantly better performance.\label{fig:4options}\label{fig2}}
\end{figure}

\begin{figure}
    \centering
    \includegraphics[width=0.8\textwidth]{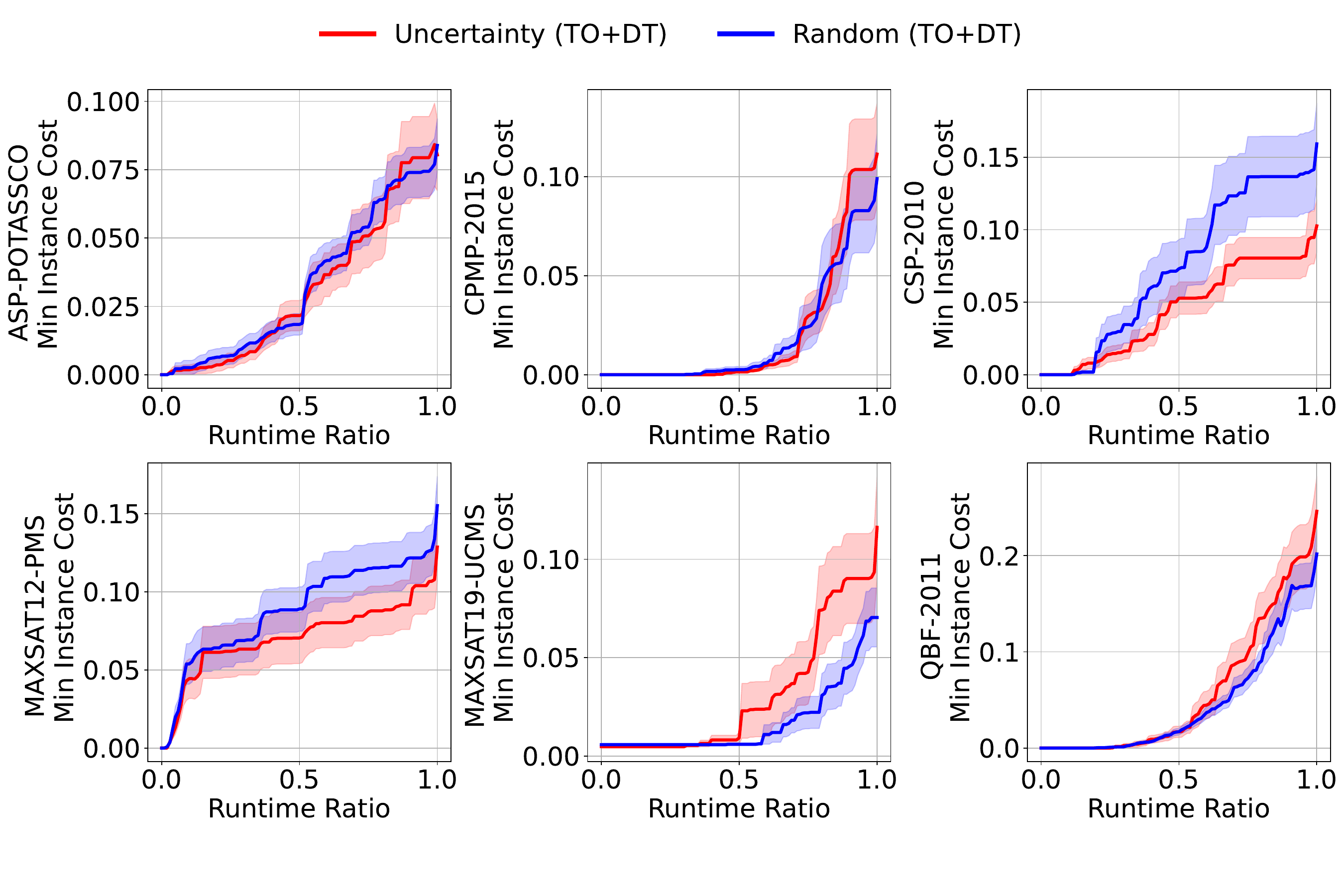}
    \vspace*{-8mm}
    \caption{TO+DT options (best in \Cref{fig:4options}) and disaggregating by approach. No clear winner.\label{fig:todt}\label{fig3}}
\end{figure}

% \vspace{-3.2 mm}

\subsection*{Key findings}

Even in the worst configuration of frugal algorithm selection, we are able to reduce the labelling cost without sacrificing predictive performance in comparison to passive learning (See \Cref{fig2}).
% , Baseline for the QBF-2011 dataset).
In several cases the training effort is reduced to 10\% of the labelling cost of passive learning. It is clear that our frugal approaches are able to reduce the training cost independent of configuration.

\Cref{fig1} presents an overview of the entire set of experiments. The results are aggregated one configuration option at a time, combining results of all configurations that share, for example, Uncertainty as the instance selection method in the first plot. Overall, instance selection strategy does not make a big difference, using timeout predictors (TO) improves performance slightly, and using dynamic timeouts (DT) improves performance significantly.

Since instance selection strategy does not make a big difference and we observe an interesting interaction between TO and DT, we aggregate over the instance selection strategy and plot 4 options in \Cref{fig2}. We see that using TO slightly improves the performance, while DT improves performance significantly. Moreover the best configurations are those that use TO and DT together.

In \Cref{fig3} we focus on the configurations that include TO+DT and compare the effect of instance selection of the this best configuration setting found in \Cref{fig2}. We further validate that there is not a clear winner among uncertainty-based instance selection and random instance selection. This finding is consistent across all combinations of strategies, as detailed in \Cref{app:b}, which includes raw data plots for all eight configurations.

The observation that uncertainty-based instance selection is not better than random may be unexpected, but selecting instances purely based on informativeness (via prediction uncertainty) does not take the cost of running an instance on a particular algorithm into account. In settings where there is a uniform cost across all candidates, uncertainty-based selection may perform better than random. In our setting, however, some samples are significantly cheaper than others. In this setting, we want to balance \textit{`bang for the buck'}: maximise how much information is gained per time spent. Therefore having explicit timeout predictors and a dynamic timeout strategy makes a more significant contribution.

\section{Related work}

% NOTE: Highly related work that's probably worth a read for our next steps (AL for varying annotation cost with a single annotator, with a good overview of related work): https://link.springer.com/article/10.1007/s10994-019-05781-7

The question of how to effectively select a representative subset of benchmark instances from a large pool for a reliable and cost-effective comparison of algorithms has been investigated across different domains, including SAT~\cite{hoos2013robust,manthey2016better,fuchs2023active}, CP~\cite{matricon2021statistical}, combinatorial optimisation~\cite{misir2021benchmark}, evolutionary computation~\cite{cenikj2022selector}, and machine learning~\cite{pereira2024optimal}. While a majority of previous work focuses on selecting a subset of instances in a static setting (i.e., all instances are chosen at once), some recent work has proposed selecting instances in an iterative fashion: Matricon et al.~\cite{matricon2021statistical} present a statistical-based method to incrementally select instances for comparing performance between two solvers, while Fuchs et al.~\cite{fuchs2023active} propose an active learning-based approach for cost-effective benchmark instance selection. 

The key difference between the instance-selection approaches mentioned above and our work is that they focus on identifying the algorithm with the best \emph{overall performance} across a given problem instance distribution, while our work focuses on the algorithm selection context, where the aim is to predict the best \emph{instance-specific} algorithm. 

The closest work to ours is Volpato and Song~\cite{volpato2019active}, where three commonly-used active learning techniques were evaluated in an AS scenario for SAT~\footnote{To our binary classification models, the three techniques are identical, as detailed in \Cref{app:a}.}. However, one drawback of their work is that they did not consider the significantly varying labelling costs among algorithms and instances, despite it being a common characteristic of SAT scenarios. Consequently, the effectiveness of active learning for instance selection was reported based on the percentage of labelled data being saved, rather than their real saving cost.

% \cite{Settles2009ActiveLL} for settings, \cite{tsou2019annotation} for a utility-and-informatiion balance technique and a good literature review, \cite{hanselle2020hybrid} for the pitfall of trying to predict the cost (runtime).
Although the majority of active learning techniques assume uniform labelling cost, there exist a  number of works on non-uniform labelling cost settings (e.g.,~\cite{Settles2009ActiveLL,tomanek2010comparison,tsou2019annotation}). A common approach in those settings is to predict the labelling cost and to strive for a balance between informativeness and the predicted cost of a new data point. We adopt a similar technique in our work, where timeout predictors are used for identifying costly (unlabelled) data points. It can be considered a ``softened'' version of the cost estimation approach, as predicting the precise runtime of algorithms in the domain of combinatorial optimisation is often known to be difficult: most prominent AS techniques focus on learning the ranking among algorithms instead of trying to predict their runtime directly~\cite{hanselle2020hybrid}.

%~\cite{volpato2019active}: experiments limited to a SAT scenario, comparing three selection strategy (margin-based, uncertainty-based, and entropy-based, but performance of AL wasn't compared with random instance selection, and annotation cost variance was not considered.  margin-based was shown as the best performing one but it might not be if annotation cost is taken into account.

%Other works on selecting representative benchmark instances: \cite{hoos2013robust} for algorithm configuration, \cite{matricon2021statistical} and  \cite{cenikj2022selector} for statistical comparison among algorithms, and \cite{dietrich2024impact} for improving generalisation performance of AS models.

\section{Conclusion}
%In this paper we have proposed and evaluated empirically a number of approaches to {\em frugal} algorithm selection, in which we take an active learning approach that attempts to reduce the cost of training by using only a subset of the training instances.

We have proposed and evaluated a number of approaches to {\em frugal} algorithm selection, an active learning approach that attempts to reduce the labelling cost by using only a subset of the training data, together with a dynamic timeout strategy that uses incomplete information about performance of algorithms in the portfolio. Our results confirm the effectiveness of the proposed approach and our analysis offers useful insights regarding the contribution of each individual component of the approach. Interestingly, the standard data selection technique in active learning does not contribute much to the overall performance, while our proposed dynamic timeout mechanism results in significant improvement in cost saving. 

In future we plan to incorporate enhancement techniques in AS into the proposed active learning framework, including the use of a pre-solving schedule~\cite{xu2008satzilla} and cost-sensitive pairwise classification AS models~\cite{xu2012satzilla2012,lindauer2015autofolio}. Other important avenues include investigating the impact of hyper-parameter tuning in the active learning setting, and developing an early-stopping mechanism to terminate the learning process once diminishing returns are observed.

%\section*{Acknowledgement}
%Cirrus

\bibliography{references}

\newpage

\appendix

\section{Appendix A: Analysis of Uncertainty Measurement Behaviours in Active Learning for Binary Classification\label{app:a}}

\begin{figure}[h]
    \centering
    \includegraphics[width=1\textwidth]{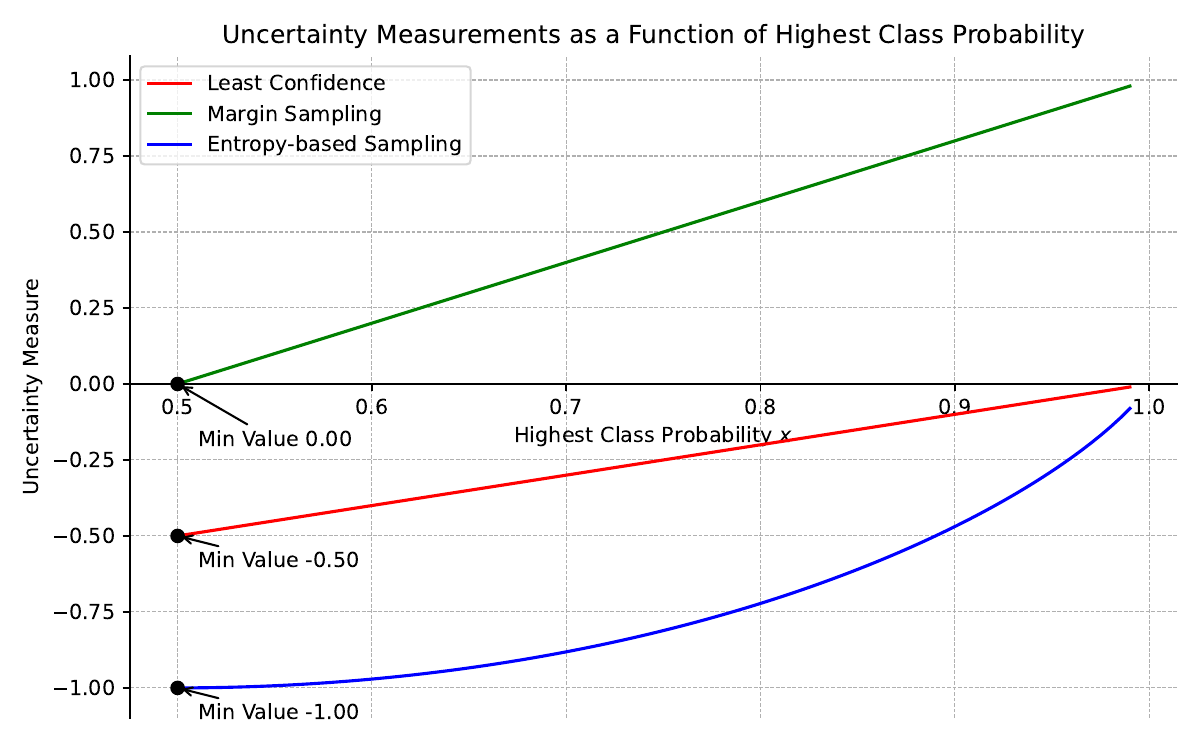}
    \caption{Uncertainty measurements as a function of the highest class probability. The red curve represents the Least Confidence uncertainty (LC) calculated as $LC = x - 1$, the green curve denotes Margin Sampling (MS) using the formula $MS = x - (1 - x)$, and the blue curve illustrates the Entropy-based method ($H(x) = -[x \log_2(x) + (1-x) \log_2(1-x)]$). Critical minimum values for each method are marked with black circles and annotated to emphasise the points where the uncertainty function is minimised.}
    \label{fig:app.1.1}
\end{figure}

There are three main approaches for uncertainty sampling in active learning. However, in a binary classification setting (which is what we use) these approaches perform identically to each other. We explain the different approaches here. \Cref{fig:app.1.1}
 shows the behaviour of these uncertainty sampling methods graphically.
 
We implement `Least Confidence` in our code.
 
\begin{itemize}
    \item \emph{Least Confidence}: for a given input $x$ and an output label $\hat{y}$, we can measure the posterior probability $P(\hat{y}|x; \theta)$ of observing $\hat{y}$ given $x$ via the current model (parameterised by $\theta$). The Least Confidence method selects data points $x^*$ with the smallest maximum posterior probability across all labels: 
    \begin{equation}
        x^* = \operatorname*{argmin}_{x} \operatorname*{max}_{\hat{y}} P(\hat{y}|x; \theta)
    \end{equation} 
    
    \item \emph{Margin-based}: this approach takes the two highest posterior probability values for each input data point $x$ and calculates their difference. The smaller the difference, the less certain the model is about its prediction and vice versa. More formally, let  $\hat{y_1}$ and $\hat{y_2}$ the output labels with the highest and second-highest posterior probabilities for a given input $x$, respectively, the queried points $x^*$ are chosen as:
    \begin{equation}
        x^* = \operatorname*{argmin}_{x} P(\hat{y_1}|x; \theta) - P(\hat{y_2}|x; \theta)
    \end{equation} 
    
    \item \emph{Entropy-based}: this approach takes into account the posterior probability values across \emph{all} output classes. The idea is to select the data points $x^*$ where there is a high entropy among the predicted output labels:
    \begin{equation}
        x^* = \operatorname*{argmax}_{x} - \sum_{i}P(\hat{y}|x; \theta) \log P(\hat{y}|x; \theta)
    \end{equation} 
\end{itemize}

\newpage

\section{Appendix B: Performance of 8 individual configurations\label{app:b}}

\begin{figure}
    \centering
    \includegraphics[width=1\textwidth]{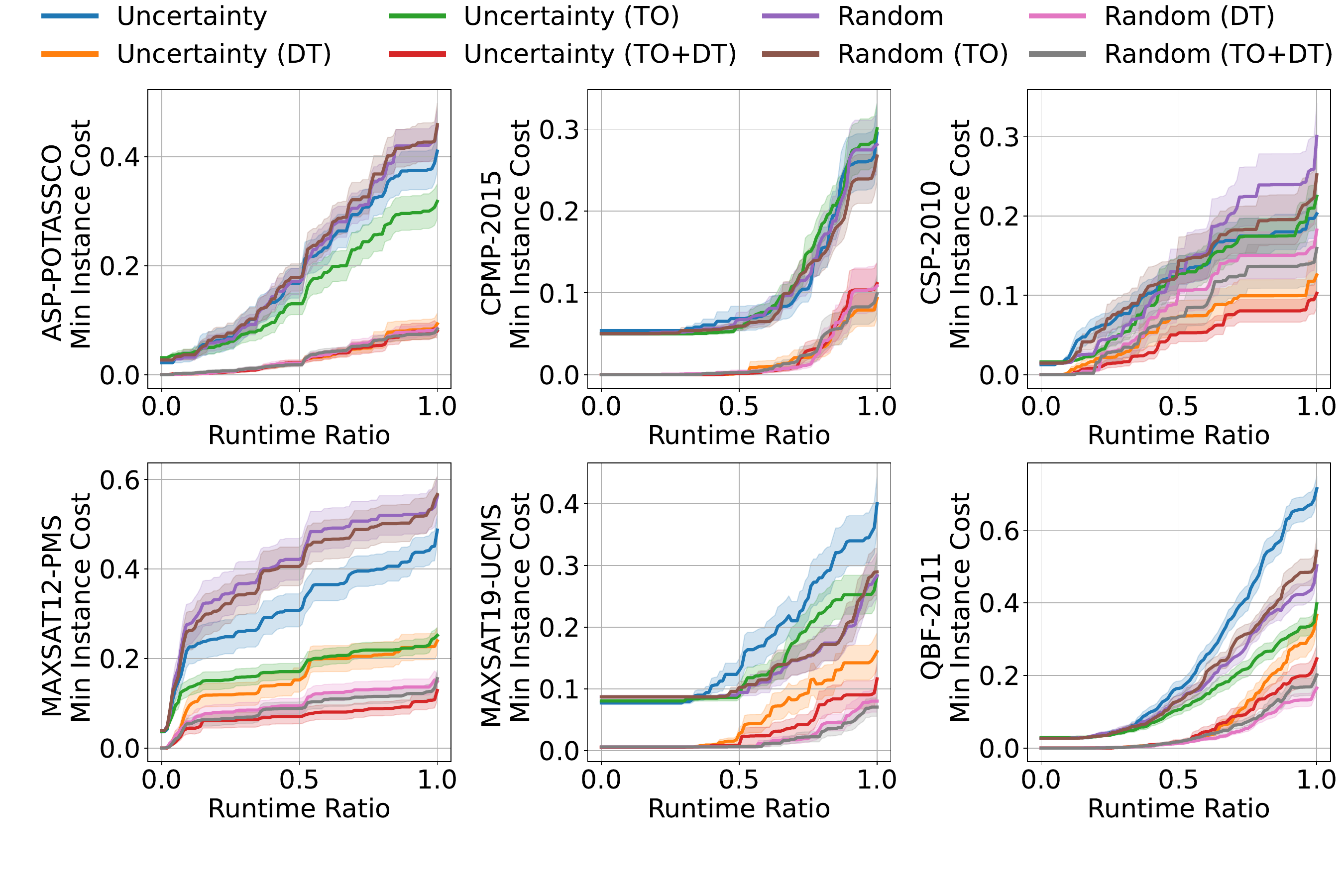}
    \caption{Comparison of performance across eight configurations as described in the paper. Each configuration was normalised according to the passive learning prediction performance ratio.}
    \label{fig:app.2.1}
\end{figure}

\Cref{fig:app.2.1} illustrates a side-by-side comparison of the following eight active learning strategies in binary classification without aggregation across configurations:

\begin{itemize}
    \item Uncertainty Sampling (Baseline)
    \item Uncertainty Sampling with Timeout Predictor (TO)
    \item Uncertainty Sampling with Dynamic Timeout (DT)
    \item Uncertainty Sampling with Timeout Predictor and Dynamic Timeout (TO+DT)
    \item Random Sampling (Baseline)
    \item Random Sampling with Timeout Predictor (TO)
    \item Random Sampling with Dynamic Timeout (DT)
    \item Random Sampling with Timeout Predictor and Dynamic Timeout (TO+DT)
\end{itemize}

% \section*{Appendix C: Experimental setup}

% In this study, we employed the Random Forest algorithm, enhanced by random bootstrapping, as the primary machine learning model. The configuration was adapted from Autofolio's hyperparameters, specifically optimised for algorithm selection tasks, with the exclusion of the min\_samples\_leaf parameter to mitigate damaging prediction performance during early phases of active learning. To validate the robustness of our study, we employed a cross-validation approach with 10 splits. For reproducibility, we utilised 5 distinct seeds (7, 42, 99, 123, 12345) in our experiments, ensuring consistent generalisation across multiple runs. In addition, the experiment includes additional parameters that include the configurations we propose:

% \textbf{timeout\_predictor\_usage} This parameter was used to decide whether the timeout predictor is used or not on the system.

% \textbf{timeout\_limit} This parameter is the parameter that specifies the initial time of the dynamic timeout. Dynamic timeout is determined as 3600 seconds in cases where it is not used

% \textbf{initial\_train\_size} This parameter specifies how many data points will be used for the initial training in active learning

% \textbf{query\_size} In our study, the query size is defined as the parameter that specifies the percentage of the dataset to be queried in each iteration. For our experiments, this was set at 1\%. 

\newpage

\section{Appendix C: Description Table of Selected Datasets\label{app:c}}
\begin{table}[ht]
\begin{adjustbox}{center}
\begin{tabular}{l|rrr|rrr}
Dataset &  Instances &  Algorithms &  Features &  Total Time & VBS &  SBS\\ \hline
ASP-POTASSCO & 1294 & 11 & 138 & 2,085h & 8h & 112h\\ 
CPMP-2015 &527 &4 &22 & 682h & 33h & 134h\\
CSP-2010 & 2024 & 2 & 86 & 435h & 49h & 82h\\
MAXSAT12-PMS & 876 & 6 & 37 & 1,472h & 8h & 85h\\
MAXSAT19-UCMS & 572 & 7 & 54 & 545h & 20h & 52h\\
QBF-2011 & 1368 & 5 & 46 & 352h & 28h & 300h\\
\end{tabular}
\end{adjustbox}
\caption{Descriptive statistics of selected datasets. Times rounded to the nearest whole number.\label{fig:app.4.1}}
\end{table}

\newpage

\section{Appendix D: Timeout (TO) Configuration Impact on Passive Learning\label{app:d}}

\begin{figure}[!ht]
    \centering
    \includegraphics[width=1\textwidth]{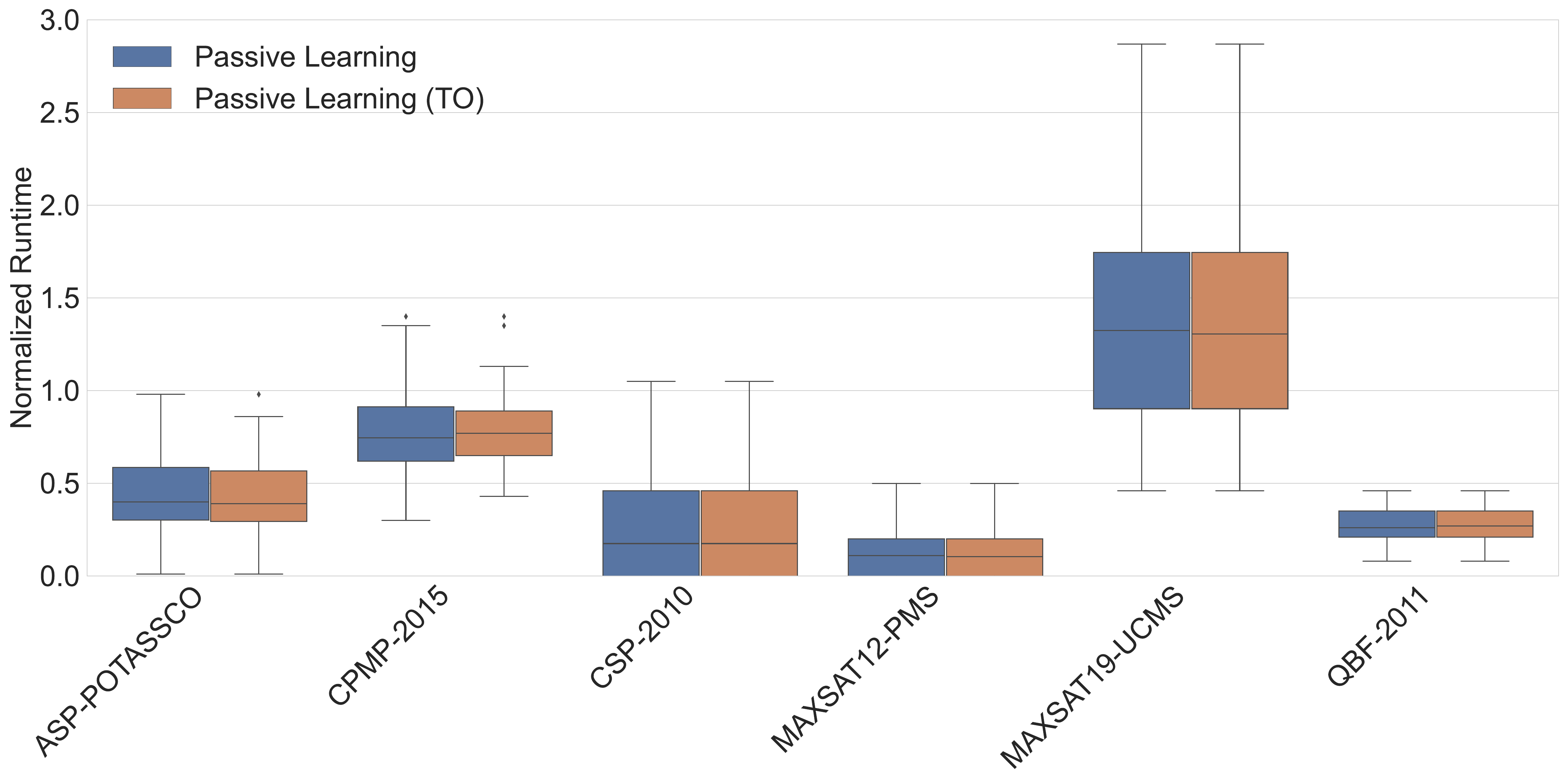}
    \caption{ Comparison of Timeout (TO) Configuration Impact on Passive Learning: The graph illustrates that implementing the TO configuration in passive learning on the test set does not significantly enhance performance, yet importantly, it does not compromise prediction accuracy either.}
    \label{fig:app:5.1}
\end{figure}

\end{document}